\documentclass[letterpaper, 10 pt, conference]{ieeeconf}  
\IEEEoverridecommandlockouts                              
\usepackage{packages}

\title{\LARGE\bf 

A Generalization of CLAP from 3D Localization to Image Processing, A Connection With RANSAC \& Hough Transforms

}
\author{Ruochen Hou$^{1}$, Gabriel I.~Fernandez$^{1}$, Alex Xu$^{1}$, and Dennis W.~Hong$^{1}$
\thanks{
$^{1}$Robotics and Mechanisms Laboratory (RoMeLa), Department of Mechanical and Aerospace Engineering, University of California, Los Angeles, CA 90095, USA.
        {\tt\small \{houruochen, gabriel808, ax90l0tl, dennishong\}@ucla.edu}
        }
}

\begin{document}
\maketitle
\thispagestyle{empty}
\pagestyle{empty}

\begin{abstract}
In previous work, we introduced a 2D localization algorithm called CLAP, Clustering to Localize Across $n$ Possibilities, which was used during our championship win in RoboCup 2024, an international autonomous humanoid soccer competition. CLAP is particularly recognized for its robustness against outliers, where clustering is employed to suppress noise and mitigate against erroneous feature matches. This clustering-based strategy provides an alternative to traditional outlier rejection schemes such as RANSAC, in which candidates are validated by reprojection error across all data points. In this paper, CLAP is extended to a more general framework beyond 2D localization, specifically to 3D localization and image stitching. We also show how CLAP, RANSAC, and Hough transforms are related. The generalization of CLAP is widely applicable to many different fields and can be a useful tool to deal with noise and uncertainty.


\end{abstract}


\section{Introduction}
\label{sec:intro}
Robust localization and perception are fundamental problems in robotics and computer vision. In real-world environments, algorithms must operate reliably despite noise, outliers, occlusions, and ambiguous observations. Traditional approaches, such as optimization-based registration or probabilistic filtering, often rely on carefully tuned models or iterative refinement to achieve accuracy. However, their performance can degrade significantly when subjected to high levels of false detections or systematic ambiguities. This motivates the need for algorithms that are intrinsically robust to outliers without relying heavily on prior assumptions or costly re-estimation.

In recent work, we introduced CLAP: Clustering to Localize Across $n$ Possibilities \cite{fernandez2025clapclusteringlocalizen}. CLAP made its first debut as a localization algorithm at RoboCup 2024, an international autonomous humanoid soccer competition where we dominantly won \cite{symposium}. The key idea of CLAP is to generate multiple candidate estimates from geometric relationships between observed features, and then apply clustering to identify the most consistent subset. Unlike outlier rejection schemes such as RANSAC, which validate hypotheses by evaluating reprojection error across all data points, CLAP exploits the natural clustering tendency of correct solutions. This property allows CLAP to remain robust even when the majority of inputs are noisy or incorrect, a characteristic that proved essential during RoboCup, a highly dynamic competitive settings.

While originally designed for 2D robot localization on a soccer field, the underlying principle of CLAP—leveraging clustering to isolate consistent solutions—extends beyond this domain. Many estimation problems in robotics and computer vision share similar structures: multiple noisy hypotheses arise from ambiguous correspondences, symmetries, or partial observations, yet the correct solutions form coherent clusters in the solution space. This suggests that CLAP is not merely a specialized localization method, but rather an instance of a broader framework for robust inference.

In this paper, we generalize the CLAP methodology into a flexible framework applicable to a wider range of tasks. We demonstrate its extension to 3D localization problems, where robots must estimate full six-degree-of-freedom poses under uncertainty, and to image processing tasks, such as robust homography estimation, where clustering of geometric hypotheses yields improved stability. By reformulating CLAP in a generalized setting, we highlight its potential as a unifying approach to outlier resistance across domains.

The main contributions of this paper are as follows:
\begin{itemize}
    \item We formalize CLAP as a general clustering-based framework for robust estimation under outliers.
    \item We extend the method to 3D localization.
    \item We explore applications of the framework in image processing tasks, image stitching, showing that clustering-based consistency checks improve robustness compared to conventional outlier rejection methods.
    \item Through these extensions, we show that the idea behind CLAP is not limited to humanoid soccer but represents a more general paradigm for robust inference in robotics and computer vision.
\end{itemize}

\section{Related Work}
\label{sec:related_work}
Robust estimation in the presence of outliers has been a long-standing research challenge in robotics and computer vision. Classical approaches such as the Random Sample Consensus (RANSAC) \cite{fischler1981random} family of algorithms address this problem by iteratively generating hypotheses from minimal subsets of data and then validating each hypothesis by evaluating reprojection errors across all available measurements. While RANSAC and its many variants have been widely adopted \cite{raguram2012usac}, their performance often depends on careful tuning of thresholds and can suffer when the fraction of outliers is high.

Another line of research focuses on consensus-based methods. These approaches exploit the principle that correct hypotheses are supported by a large, consistent subset of the data, while incorrect hypotheses fail to accumulate consensus. The Hough transform \cite{hough1962method} is one of the earliest and most influential examples: it maps feature detections into a parameter space where correct geometric models, such as lines or circles, manifest as dense clusters of votes. In this sense, the Hough transform can be interpreted as a special case of clustering-based inference, where accumulation in the parameter space serves as the clustering mechanism.

The Clustering to Localize Across n Possibilities (CLAP) method builds upon this consensus-based philosophy. Instead of validating hypotheses individually, CLAP generates a set of possible states from geometric feature pairs and identifies the most consistent solutions through clustering in the state space. This clustering perspective distinguishes CLAP from RANSAC-style hypothesis testing: rather than discarding incorrect estimates through iterative reprojection checks, CLAP leverages the natural density structure of correct solutions to suppress outliers.

Related paradigms include density-based clustering methods (e.g., DBSCAN), model-based clustering (e.g., Gaussian Mixture Models), and mean-shift-based techniques, which have all been applied in robust vision and robotics tasks. More recently, learning-based approaches have attempted to learn robust representations directly from data; however, these often require extensive training datasets and may generalize poorly to unseen environments.

By unifying these perspectives, we view CLAP not as an isolated method but as part of a broader family of clustering-driven robust estimation techniques. In particular, its connection to both RANSAC and the Hough transform highlights that clustering in solution or parameter space can serve as a powerful alternative to explicit hypothesis rejection. This insight motivates the generalization of CLAP beyond its original 2D localization application, enabling extensions to 3D localization and image processing tasks.
\section{Generalized CLAP Framework}
\label{sec:Generalized_CLAP_Framework}

The Clustering to Localize Across $n$ Possibilities (CLAP) framework is a general strategy for robust estimation in the presence of outliers. The key idea is that correct hypotheses form dense clusters in an appropriate space, while incorrect hypotheses remain scattered. By identifying and averaging the cluster center, CLAP yields estimates that are robust against noise and spurious matches.

\subsection{Spaces for Clustering}
Depending on the problem setting, CLAP can be applied in different spaces:
\begin{itemize}
    \item \textbf{Input space:} clustering is performed directly on raw sensor observations. For example, landmark detections or feature descriptors can be grouped to filter noisy measurements before model inference.
    \item \textbf{State (output) space:} clustering is performed on the estimated states of the system. In localization, each hypothesis corresponds to a candidate robot pose in $SE(2)$ or $SE(3)$, and clustering identifies the most consistent pose cluster. The state space coincides with the final output to be estimated.
    \item \textbf{Parameter/model space:} clustering is performed on candidate models explaining the data. For example, in image stitching, homographies $H \in PGL(3)$ are clustered to obtain a stable alignment. Classical methods such as the Hough transform can also be interpreted as CLAP in parameter space, since peaks represent dense clusters of consistent model parameters.
\end{itemize}
This generality highlights CLAP as a unifying principle: the choice of space depends on whether one seeks to cluster observations, system states, or model parameters.

\subsection{Distance Metrics and Averaging}
To perform clustering, a suitable distance metric must be defined in the chosen space. In state spaces such as $SE(3)$, this may involve Lie-algebra distances or relative transform errors. In parameter spaces such as homographies, Frobenius or geodesic distances can be applied. Once distances are defined, clusters can be identified using neighborhood filtering, density estimation, or robust center-finding methods. The final estimate is obtained by averaging, for example via medoid selection, Karcher mean, or log--Euclidean mean.

\subsection{Local vs. Global Clustering Modes}
An important design choice in CLAP is whether clustering is performed locally or globally:
\begin{itemize}
    \item \textbf{Local clustering} restricts hypotheses to a neighborhood around the previous estimate. This improves efficiency and ensures smooth updates, making it well-suited for online real-time applications with frequent updates.
    \item \textbf{Global clustering} considers all hypotheses across the entire space. This mode provides resilience against catastrophic errors and allows recovery from large deviations, at the cost of greater computational effort.
\end{itemize}
In practice, these modes can be combined: local clustering is used for efficiency in nominal operation, while global clustering is invoked periodically or when failures are detected.

\subsection{Summary}
The CLAP framework thus consists of three core components: (1) generating multiple candidate solutions, (2) defining distances to enable clustering in the relevant space, and (3) computing a robust cluster center as the final estimate. By decoupling estimation from outlier rejection, CLAP generalizes across domains, from robot localization in $SE(3)$ to image stitching via homography clustering.

\section{Application I: Extension to 3D Localization}
\label{sec:3D_Localization}

Localization in 3D environments requires estimating the full pose of a robot, represented as a transformation $T \in SE(3)$. Extending CLAP from 2D to 3D follows the same principle: generate multiple pose hypotheses from feature correspondences, then use clustering to identify the most consistent solutions.  

\subsection{Candidate Pose Generation}
\begin{figure}[htbp]
    \centering
    \begin{subfigure}{0.235\textwidth}
        \includegraphics[width=\linewidth]{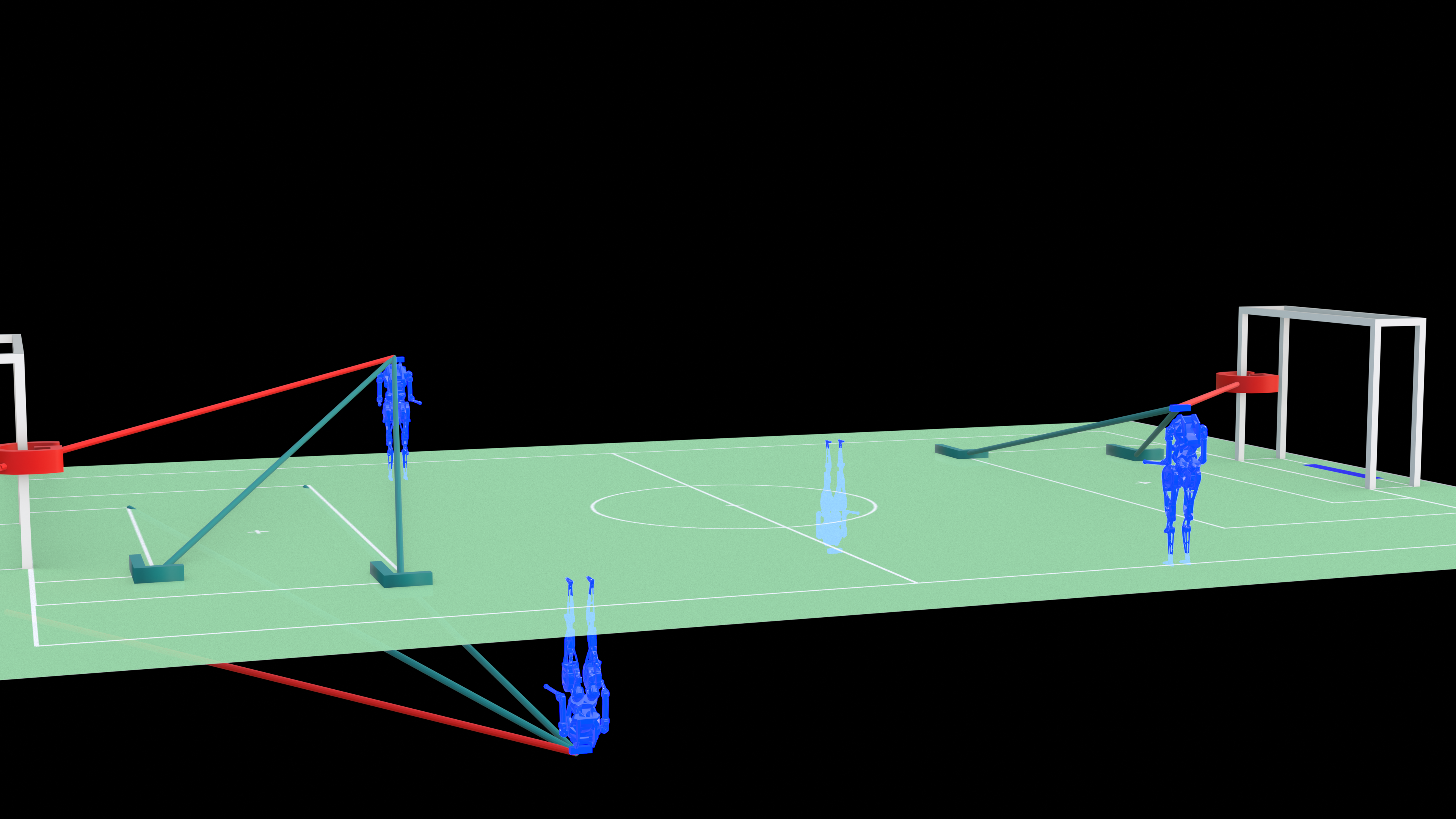}
        \caption{G-L-L Landmarks}
        \label{fig:GLL}
    \end{subfigure}
    \begin{subfigure}{0.235\textwidth}
        \includegraphics[width=\linewidth]{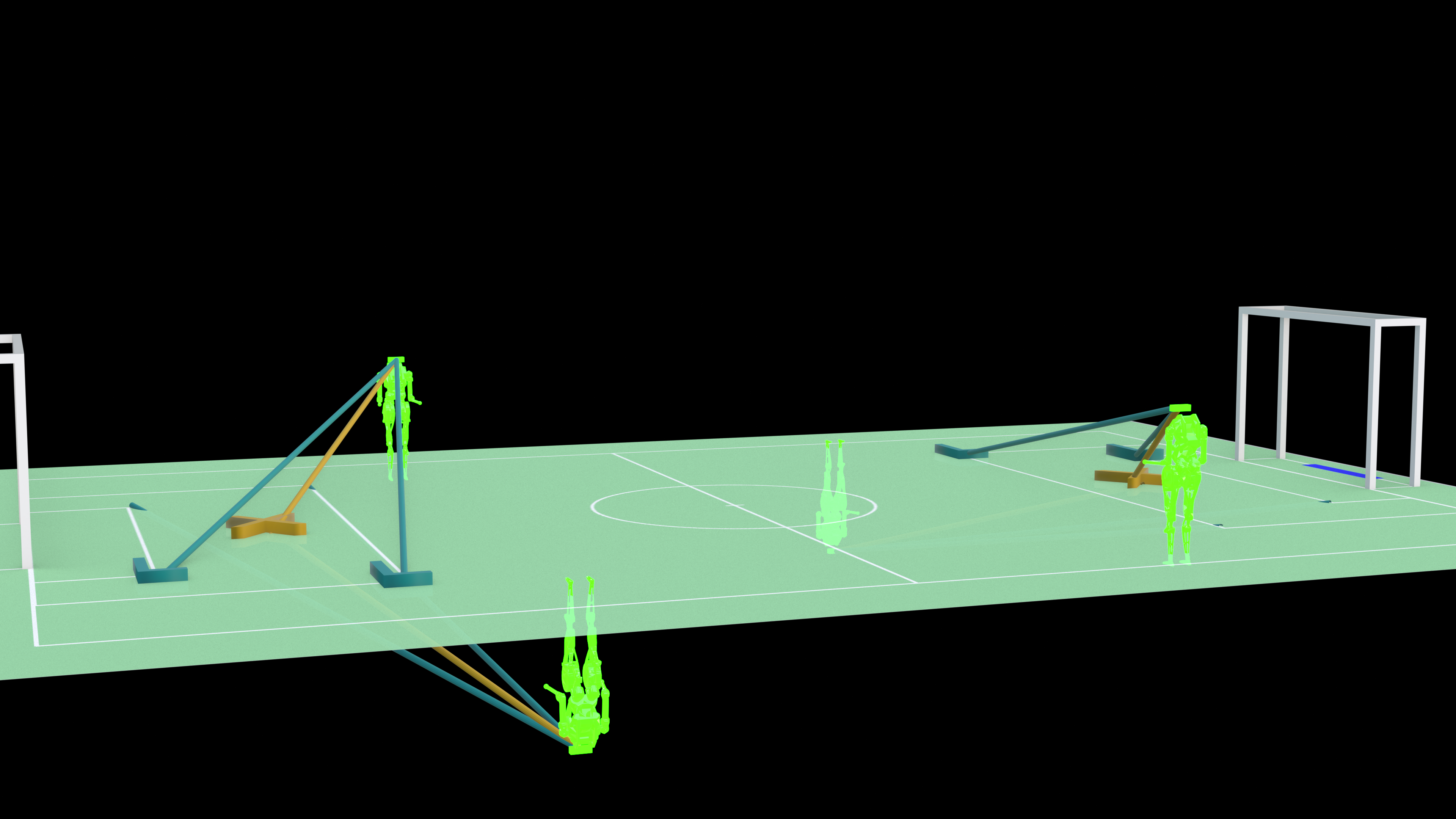}
        \caption{L-L-X Landmarks}
        \label{fig:LLX}
    \end{subfigure}
    
    \begin{subfigure}{0.235\textwidth}
        \includegraphics[width=\linewidth]{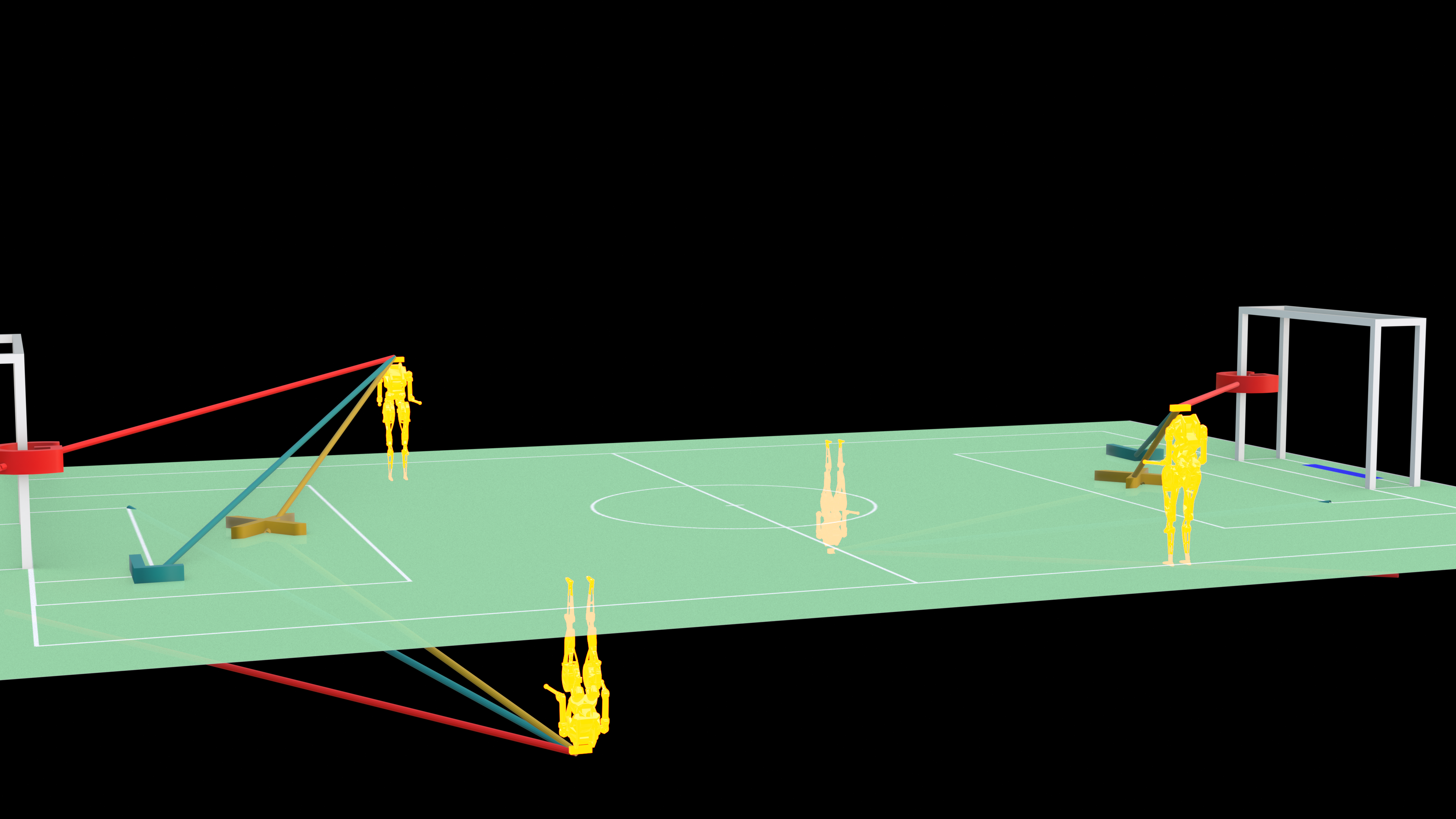}
        \caption{G-X-L Landmarks}
        \label{fig:GXL}
    \end{subfigure}
    \begin{subfigure}{0.235\textwidth}
        \includegraphics[width=\linewidth]{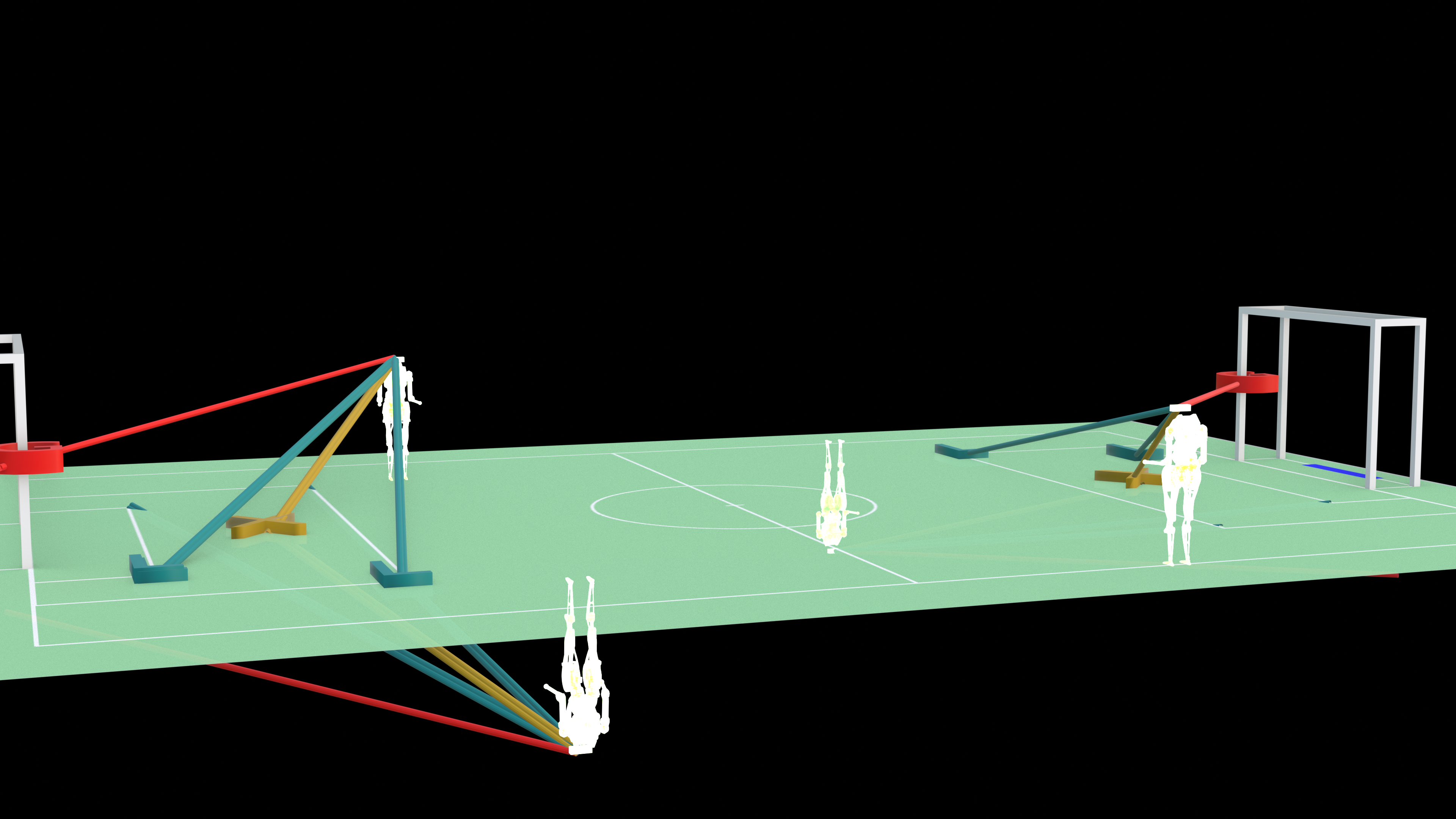}
        \caption{Combined}
        \label{fig:combined}
    \end{subfigure}
    
    \caption{Visualization of 4 landmark observations split into three subsets. Each set of three landmarks has 4 possible positions. When stacking all of these estimates, the combined estimate is white due to the mixing of colors.}
    \label{fig:local_ex}
\end{figure}

Candidate poses are generated by solving minimal problems with subsets of observed landmarks. In our implementation, triplets of landmarks are matched to the global map, and each correspondence yields a candidate pose. An example is shown in \cref{fig:local_ex} where a set of four landmarks are observed: \textit{G}, \textit{L}, \textit{X}, and \textit{L}. Three subsets are shown with their corresponding poses in \cref{fig:GLL,fig:LLX,fig:GXL}, and \cref{fig:combined} shows the result of combining each estimate. The pose solver employs a closed-form SVD-based alignment method~
\cite{kabsch1976solution,arun1987least,maurer1996registration}
, ensuring efficient and numerically stable estimation even under noise. This results in a large set of possible poses, many of which are incorrect due to ambiguous correspondences or outliers.  

\subsection{Distance Metrics on $SE(3)$}
To cluster poses, we define distances between transformations. Several complementary metrics are used:  

\begin{enumerate}
    \item \textbf{Relative Transform Error:}  
    \[
    \Delta = T_1^{-1} T_2, \quad 
    (e_t, e_r) = (\|t_\Delta\|, \|\log(R_\Delta)\|),
    \]  
    measuring translational and rotational errors separately.  

    \item \textbf{Lie-Log Scalar Distance:}  
    Using the logarithm map on $SE(3)$, the error twist $\xi = \log(T_1^{-1} T_2)$ defines  
    \[
    d_{SE(3)} = \sqrt{\|\rho\|^2 + \lambda^2 \|\phi\|^2},
    \]  
    where $\rho$ and $\phi$ are translational and rotational components.  

    \item \textbf{Point-Set Induced Distance:}  
    For a set of reference points $P$, the distance compares transformed point clouds under two candidate poses, capturing both translational and rotational misalignments.  
\end{enumerate}

These metrics allow flexible filtering, depending on whether translation accuracy, rotation accuracy, or task-specific consistency is more critical.  

\subsection{Pose Filtering and Clustering}
Given an initial guess $T_{\text{init}}$, candidate poses are filtered by thresholding their distance to the reference pose. Poses within the tolerance are retained and ranked. This step is analogous to local clustering around the initial guess, discarding outliers that are too far in either translation or rotation.  

\subsection{Averaging and Final Estimate}
The surviving poses form a dense cluster in $SE(3)$. We compute the cluster center using one of several averaging schemes:  

\begin{itemize}
    \item \textbf{Karcher (Fréchet) Mean:} Iteratively refines the intrinsic mean in $SE(3)$.  
    \item \textbf{Log--Euclidean Mean:} Averages logarithms of poses in $\mathfrak{se}(3)$, then exponentiates.  
    \item \textbf{Split Mean:} Averages rotation and translation separately.  
\end{itemize}

The final pose estimate is taken as the averaged cluster center, ensuring robustness against outliers while maintaining smoothness across time.  

\subsection{Discussion}
This 3D extension of CLAP demonstrates the flexibility of the framework: hypotheses are generated geometrically, clustered using task-appropriate distances, and aggregated into a single robust pose estimate. Unlike particle filters or optimization-based methods, CLAP in 3D requires no iterative global search, making it suitable for real-time applications in robotics.  

\section{Application II: Image Stitching via Homography Clustering}
\label{sec:Image_Stitching}

Image stitching requires estimating a homography $H \in PGL(3)$ that maps points from one image to another. Classical pipelines rely on feature extraction, descriptor matching, and RANSAC-based filtering of correspondences. While effective, RANSAC may discard too many matches in the presence of repeated textures, or fail when outliers dominate. Extending CLAP to this problem allows robust inference without explicit per-hypothesis rejection.  

\subsection{Candidate Homography Generation}
Given matched feature points between two images, we generate candidate homographies using the Direct Linear Transform (DLT) on randomly sampled subsets of four correspondences. Thousands of such candidates are computed, each representing a plausible alignment. Many are invalid or inconsistent due to noisy matches, but the correct homographies tend to cluster in parameter space.  

\subsection{Distance Metrics on Homographies}
To enable clustering, we define pairwise distances among candidate homographies:  

\begin{itemize}
    \item \textbf{Lie-Algebra Distance:}  
    \[
    d(H_1,H_2) = \|\log(H_1^{-1}H_2)\|_F,
    \]  
    which measures geodesic deviation in the projective group.  

    \item \textbf{Frobenius Distance:}  
    A simpler alternative comparing normalized matrices.  
\end{itemize}

These metrics capture geometric consistency among candidate models, enabling robust clustering in homography space.  

\subsection{Robust Center Estimation}
Rather than searching for a single best hypothesis, CLAP identifies the cluster center as the representative solution. We implement several robust center estimators:  

\begin{itemize}
    \item \textbf{Medoid:} the candidate minimizing the sum of distances to all others.  
    \item \textbf{Lie Mean / Lie Median:} intrinsic averaging or geometric median on the Lie algebra.  
    \item \textbf{Robust Single-Center (Trimming / MAD):} iteratively refines the center by discarding outliers based on fractional trimming or median absolute deviation.  
\end{itemize}

This center-finding process replaces RANSAC’s hypothesis validation with density-driven inference, yielding stability even when many matches are ambiguous.  

\subsection{Refinement and Stitching}
Once a robust center homography is selected, it is optionally refined using inlier re-estimation with reprojection error checks. The resulting transformation is then used to warp the right image into the left image’s coordinate frame, followed by compositing. Multiple blending strategies are supported, including no-blend overlays, overwrite compositing, and distance-transform feathering for seamless panoramas.  

\subsection{Discussion}
The CLAP-based stitching approach demonstrates that clustering in parameter space is a powerful alternative to per-hypothesis rejection. Unlike RANSAC, which selects a single inlier-consistent model, CLAP aggregates information across many plausible hypotheses to recover a stable center. This makes it particularly effective in scenes with high outlier ratios, repetitive textures, or ambiguous correspondences.  

\section{Results} 
\label{sec:experiment_results}

\section{Experiments and Results}

We evaluate the generalized CLAP framework in two representative domains: 3D localization and image stitching. In both cases, we compare CLAP against baseline methods such as RANSAC and demonstrate its robustness in the presence of high outlier ratios and ambiguous correspondences.

\subsection{3D Localization Results}
To validate the 3D extension of CLAP, we use the same experimental setup as in our previous work~\cite{fernandez2025clapclusteringlocalizen}, but extend the evaluation to full six-degree-of-freedom poses. Data were collected using a motion capture (MoCap) system and a synchronized databag of landmark observations.  

Candidate poses were generated from landmark triplets and clustered in $SE(3)$ as described in Section~III. The ground truth pose from the MoCap system was used to compute translational and rotational errors.  

\begin{figure}[t]
    \centering
    \includegraphics[width=0.45\textwidth]{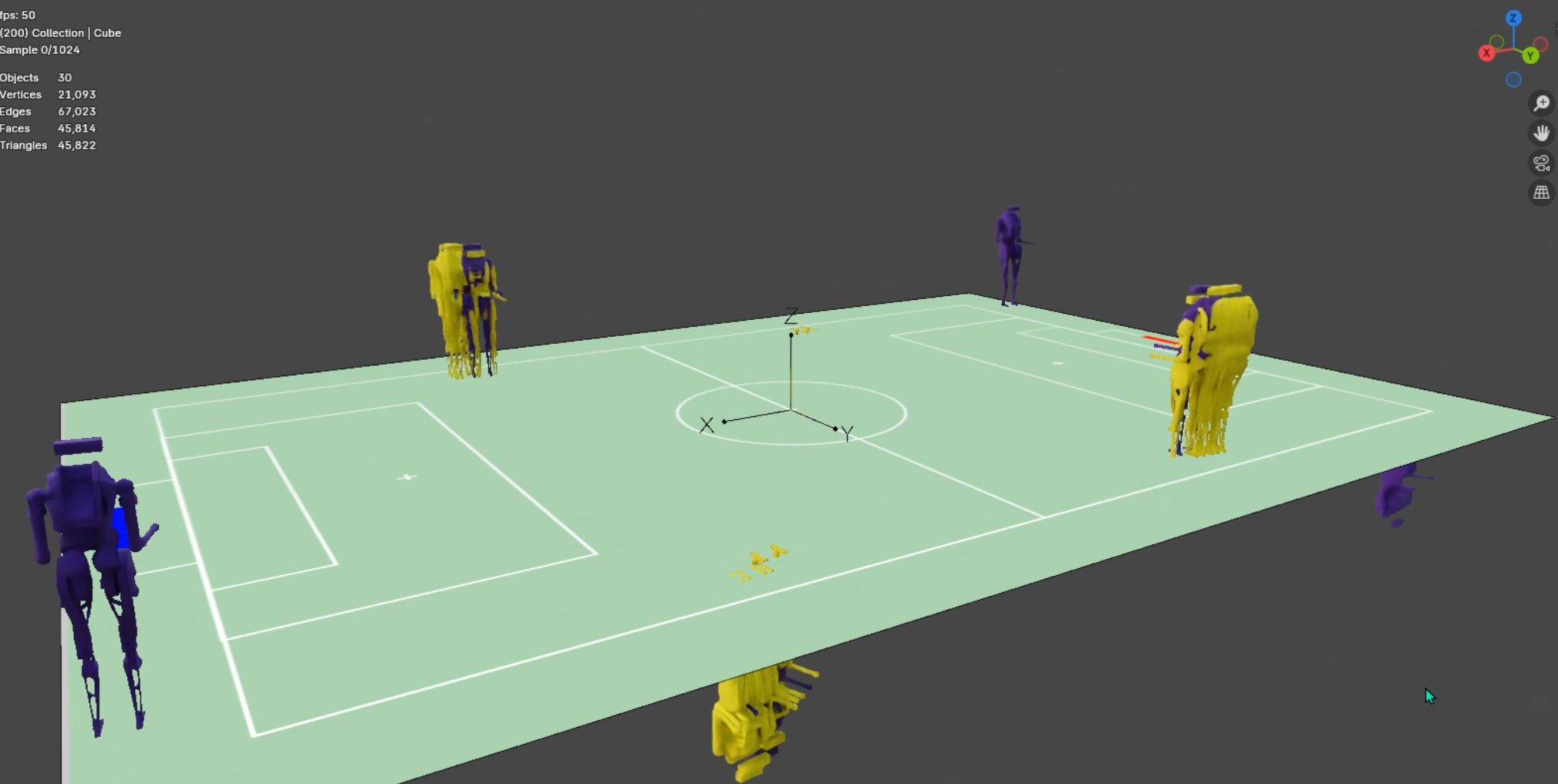}
    \caption{3D localization demo for one pose}
    \label{fig:3d_loc_error}
\end{figure}

Figure~\ref{fig:3d_loc_error} shows the cumulative distribution of localization errors. CLAP significantly outperforms RANSAC, maintaining robustness even when more than 50\% of the landmark correspondences are corrupted by outliers. The error histograms demonstrate that CLAP consistently identifies the correct cluster of poses, whereas RANSAC occasionally converges to spurious solutions when the correct inlier set is overwhelmed.  

\subsection{Image Stitching Results}
We further evaluate CLAP in the context of image stitching using the Vanishing Point Guided Stitching (VPG) dataset \cite{chen2020vanishing}. This dataset provides 36 image sets, of which S01--S12 contain ground truth matching correspondences, while S13--S36 are real-world examples without ground truth.




To establish a baseline comparison, we implement the classical DLT+RANSAC pipeline. 
For fairness, we fix the RANSAC iteration budget to 1000 and refine the estimated homography $H_{\text{RANSAC}}$ 
using the inlier set. For the CLAP method, we generate 400 candidate homographies, perform iterative clustering 
for 5 rounds, and discard 20\% of the outliers at each iteration. 
Using a fixed number of iterations makes the comparison fairer than adaptive stopping criteria, 
since both methods operate under controlled sampling budgets.

\begin{figure}[t]
    \centering
    \includegraphics[width=0.5\textwidth]{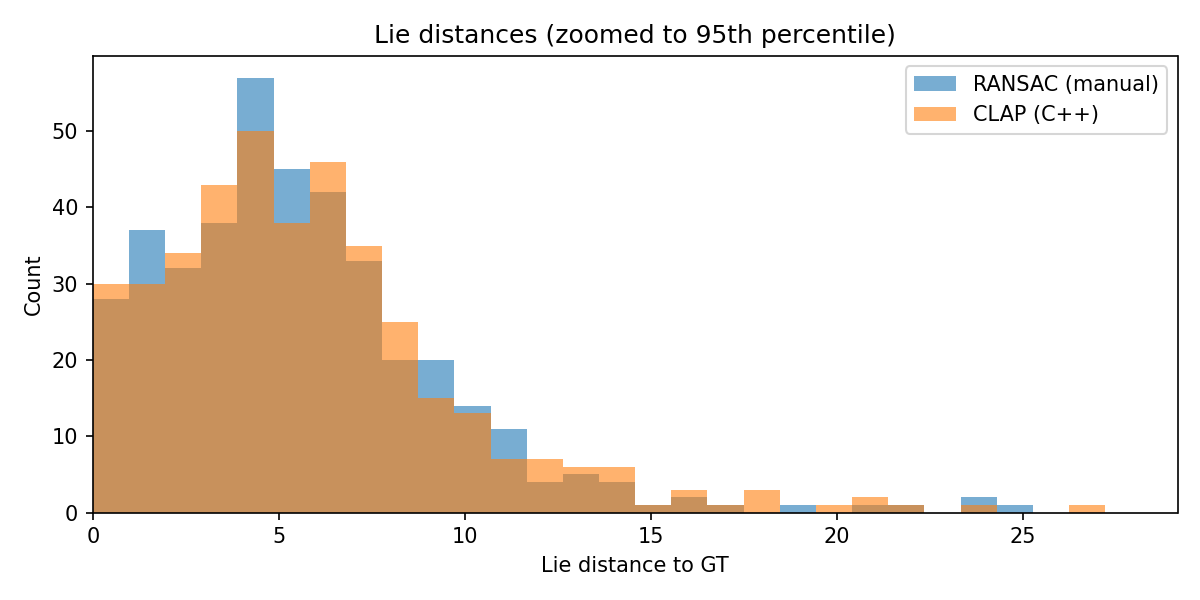}
    \caption{Lie distance to ground truth (linear scale). 
We compute $d(H,H_{\text{gt}})=\lVert \log(H_{\text{gt}}^{-1}H)\rVert_F$ using the principal matrix logarithm after normalizing $H$ so that $H_{33}=1$. 
Lower values indicate homographies closer to the ground truth.}

    \label{fig:hist_lie_distance_zoom}
\end{figure}

\begin{figure}[t]
    \centering
    \includegraphics[width=0.5\textwidth]{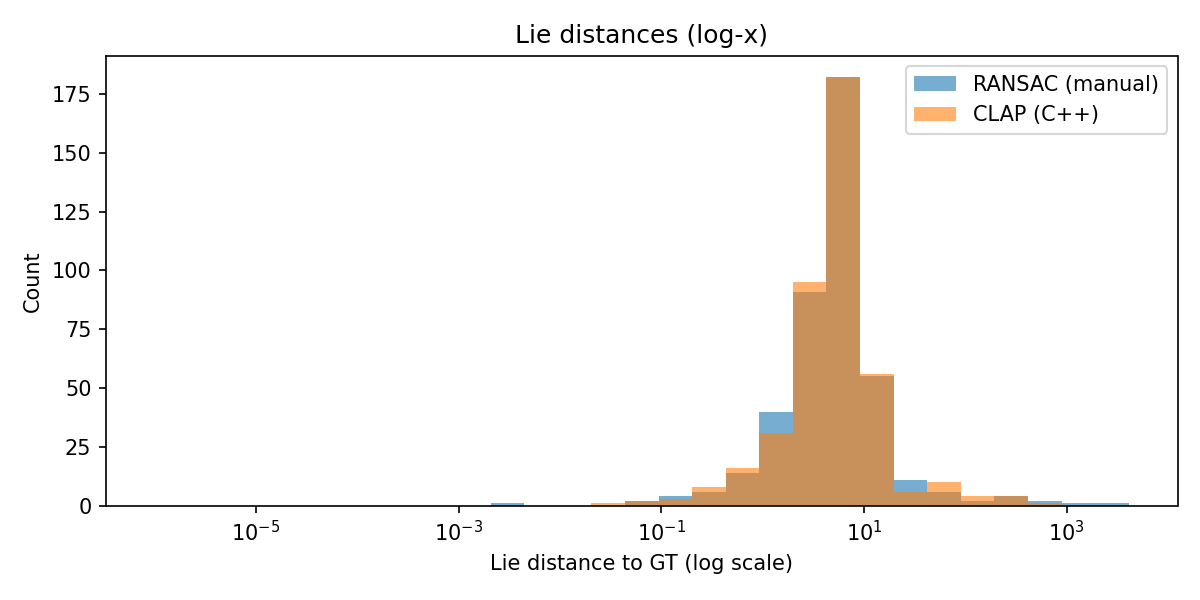}
    \caption{Lie distance to ground truth (log scale). 
Same metric as \cref{fig:hist_lie_distance_zoom}, but shown with a logarithmic x-axis (log-spaced bins with a small $\varepsilon$ offset) to highlight the heavy-tail distribution of large errors.}

    \label{fig:hist_lie_distance_log}
\end{figure}

\begin{figure}[t]
    \centering
    \includegraphics[width=0.5\textwidth]{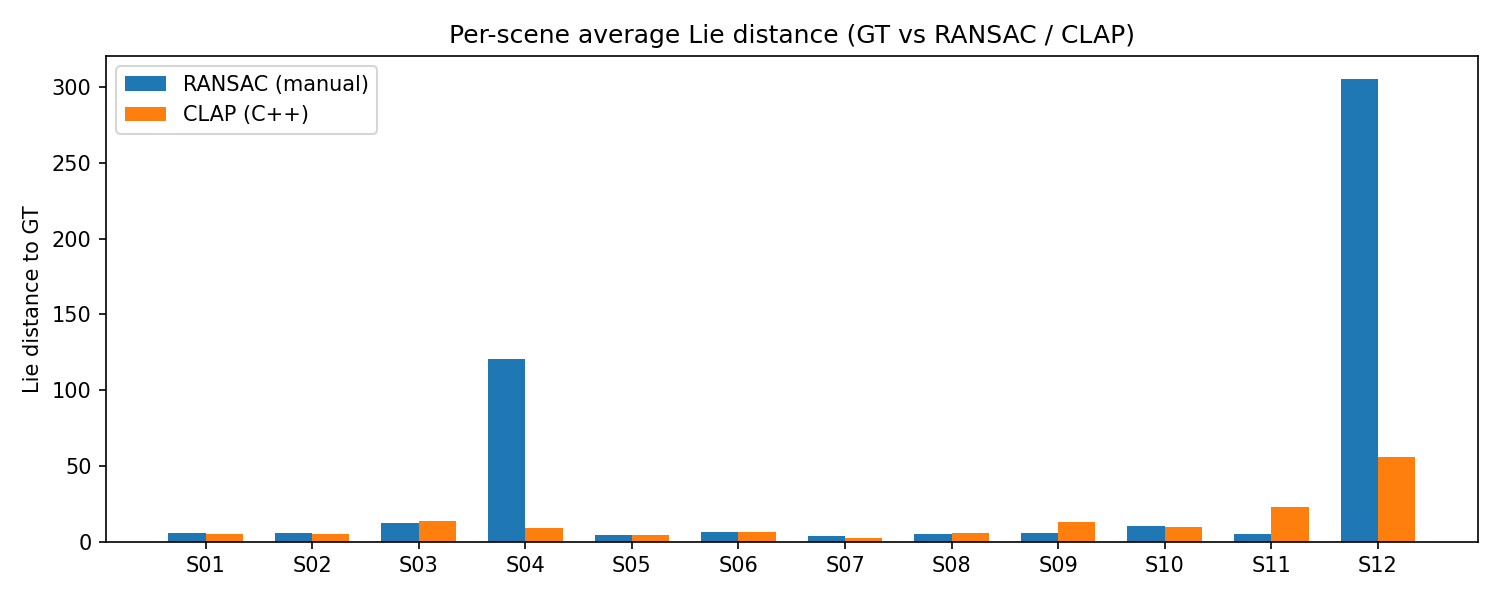}
    \caption{Per-scene Lie distance to ground truth. 
Each bar corresponds to the average Lie distance $d(H,H_{\text{gt}})=\lVert \log(H_{\text{gt}}^{-1}H)\rVert_F$ computed within one scene. 
This highlights scene-dependent performance differences, where smaller values indicate closer alignment to the ground truth homography.}

    \label{fig:per_scene_lie_distance}
\end{figure}

For the simulated subset (S01--S12), ground-truth correspondences allow us to compute a reference homography $H_{\text{GT}}$. 
We then evaluate the accuracy of $H_{\text{RANSAC}}$ and $H_{\text{CLAP}}$ by measuring their Lie distance to $H_{\text{GT}}$, 
as shown in \Cref{fig:hist_lie_distance_zoom,fig:hist_lie_distance_log,fig:per_scene_lie_distance}. 
These figures demonstrate both the global error distribution and per-scene performance, where lower values 
indicate homographies closer to the reference homography using ground truth matching. 

\begin{figure}[t]
    \centering
    \includegraphics[width=0.5\textwidth]{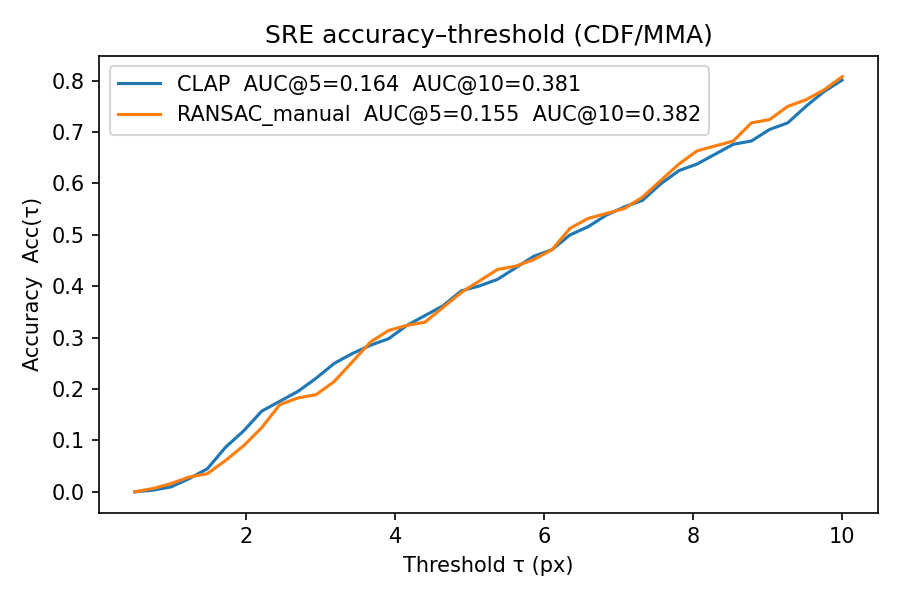}
    \caption{Cumulative distribution function (CDF) of symmetric reprojection error (SRE). 
For each method, we plot the fraction of test pairs whose SRE falls below a threshold. 
Curves shifted toward the left/top indicate more accurate and consistent homography estimation.}

    \label{fig:cdf_sre}
\end{figure}

For the real-world subset (S13--S36), ground truth matching is unavailable. Instead, we evaluate performance using the 
symmetric reprojection error (SRE) and the inlier ratio. 
\Cref{fig:cdf_sre} reports the cumulative distribution of SRE values across test pairs. 
Curves shifted toward the top-left indicate higher accuracy and consistency. The two methods show similar accuracy.

\begin{figure*}[t]
    \centering
    \includegraphics[width=1.0\textwidth]{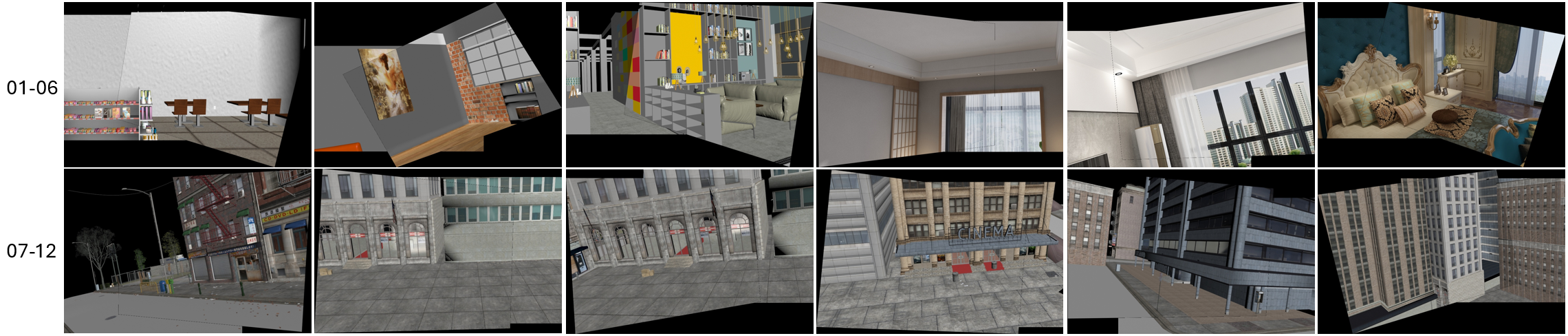}
    \caption{An overview of the CLAP image stitching results on simulation images from the VPG dataset (S01--S12). }
    \label{fig:overview_vpg_result_01_12}
\end{figure*}

\begin{figure*}[t]
    \centering
    \includegraphics[width=1.0\textwidth]{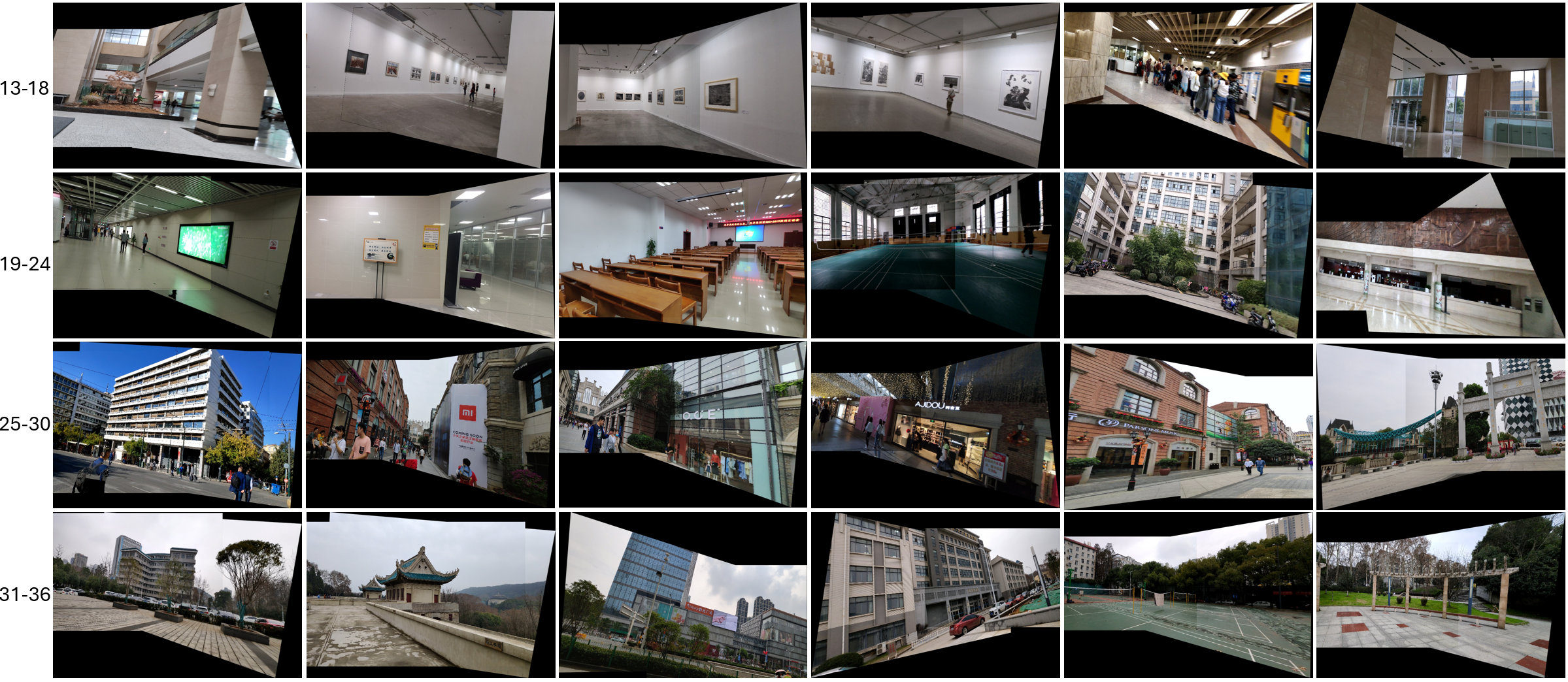}
    \caption{An overview of the CLAP image stitching results on real world images from the VPG dataset (S13--S36). }
    \label{fig:overview_vpg_result_13_36}
\end{figure*}

\subsection{Discussion}
These experiments confirm that CLAP’s clustering-based approach provides robustness beyond that of classical RANSAC. In 3D localization, clustering in $SE(3)$ reliably identifies the correct pose cluster, while in image stitching, clustering in homography space yields stable transformations even in ambiguous scenes. Together, these results demonstrate the generality of the CLAP framework across state estimation and geometric vision tasks.

\section{Discussion} 
\label{sec:Discussion}
\section{Conclusion} 
\label{sec:conclusion}

In this paper, we presented a generalization of the Clustering to Localize Across $n$ Possibilities (CLAP) framework. Originally developed for 2D localization, CLAP is based on the principle that correct hypotheses form consistent clusters, while outliers remain dispersed. By abstracting this idea, we demonstrated that CLAP can be applied in multiple domains by clustering candidates in different spaces, including input, state (output), and parameter/model spaces.  

We instantiated this generalized framework in two applications. First, we extended CLAP to 3D localization, where candidate robot poses in $SE(3)$ are clustered and averaged to obtain robust six-degree-of-freedom pose estimates. Second, we applied CLAP to image stitching, where candidate homographies are clustered in projective space, providing a stable alternative to RANSAC-based methods for robust panorama generation.  

These results highlight CLAP as a unifying paradigm for robust estimation across robotics and vision tasks. Rather than discarding outliers one by one, CLAP exploits the density structure of correct solutions, offering robustness in scenarios with high noise, ambiguous correspondences, and repeated patterns.  

\subsection*{Future Work}
Future directions include extending the framework to other domains such as multi-view geometry, multi-robot localization, and sensor fusion, where clustering across heterogeneous candidate spaces may further enhance robustness. In addition, integrating learning-based priors with CLAP could enable hybrid methods that combine data-driven feature selection with clustering-driven robustness.



\end{document}